\documentclass[10pt,twocolumn,letterpaper]{article}

\usepackage{iccv}              

\definecolor{iccvblue}{rgb}{0.21,0.49,0.74}
\usepackage[pagebackref,breaklinks,colorlinks,allcolors=iccvblue]
{hyperref}

\usepackage{amsmath}
\usepackage{multirow}
\usepackage{float}
\usepackage{xcolor}   
\usepackage{amsmath}   
\usepackage{siunitx} 
\usepackage{marvosym}

\def\paperID{7756} 
\def\confName{ICCV}
\def\confYear{2025}

\title{ADCD-Net: Robust Document Image Forgery Localization via \\ Adaptive DCT Feature and Hierarchical Content Disentanglement}
\author{
Kahim Wong\(^{1}\), Jicheng Zhou\(^{1}\), Haiwei Wu\(^{2}\), Yain-Whar Si\(^{1}\), Jiantao Zhou\(^{1}\)\Letter \\
{\small \(^{1}\)State Key Laboratory of Internet of Things for Smart City, Department of Computer and Information Science, University of Macau}\\
{\small\(^{2}\)School of Information and Software Engineering, University of Electronic Science and Technology of China}\\
{\tt\small \{yc37437, mc35093, fstasp, jtzhou\}@um.edu.mo, haiweiwu@uestc.edu.cn}
}

\begin{document}
\maketitle

\newcommand\blfootnote[1]{%
  \begingroup
  \renewcommand\thefootnote{}\footnote{#1}%
  \addtocounter{footnote}{-1}%
  \endgroup
}

\begin{abstract}

The advancement of image editing tools has enabled malicious manipulation of sensitive document images, underscoring the need for robust document image forgery detection.\blfootnote{\Letter \ Corresponding author} Though forgery detectors for natural images have been extensively studied, they struggle with document images, as the tampered regions can be seamlessly blended into the uniform document background (BG) and structured text. On the other hand, existing document-specific methods lack sufficient robustness against various degradations, which limits their practical deployment. This paper presents ADCD-Net, a robust document forgery localization model that adaptively leverages the RGB/DCT forensic traces and integrates key characteristics of document images. Specifically, to address the DCT traces' sensitivity to block misalignment, we adaptively modulate the DCT feature contribution based on a predicted alignment score, resulting in much improved resilience to various distortions, including resizing and cropping. Also, a hierarchical content disentanglement approach is proposed to boost the localization performance via mitigating the text-BG disparities. Furthermore, noticing the predominantly pristine nature of BG regions, we construct a pristine prototype capturing traces of untampered regions, and eventually enhance both the localization accuracy and robustness. Our proposed ADCD-Net demonstrates superior forgery localization performance, consistently outperforming state-of-the-art methods by 20.79\% averaged over 5 types of distortions. The code is available at \url{https://github.com/KAHIMWONG/ACDC-Net}.

\end{abstract}

\section{Introduction}
With the rise of image editing tools such as Photoshop, Canva, and deep inpainting models~\cite{9537606, yang2020swaptext}, altering document images has become effortless. This allows adversaries to manipulate financial records or other sensitive data in the document images to conceal malicious activities such as spreading rumors and committing economic fraud, which can lead to costly errors in decision-making and significant financial loss~\cite{verdoliva2020media}. These manipulations pose significant security threats, making effective tools for detecting tampering in document images more critical than ever~\cite{qu2023towards, dong2024robust}.

\begin{figure}[t]
    \centering
    \includegraphics[width=0.46\textwidth]{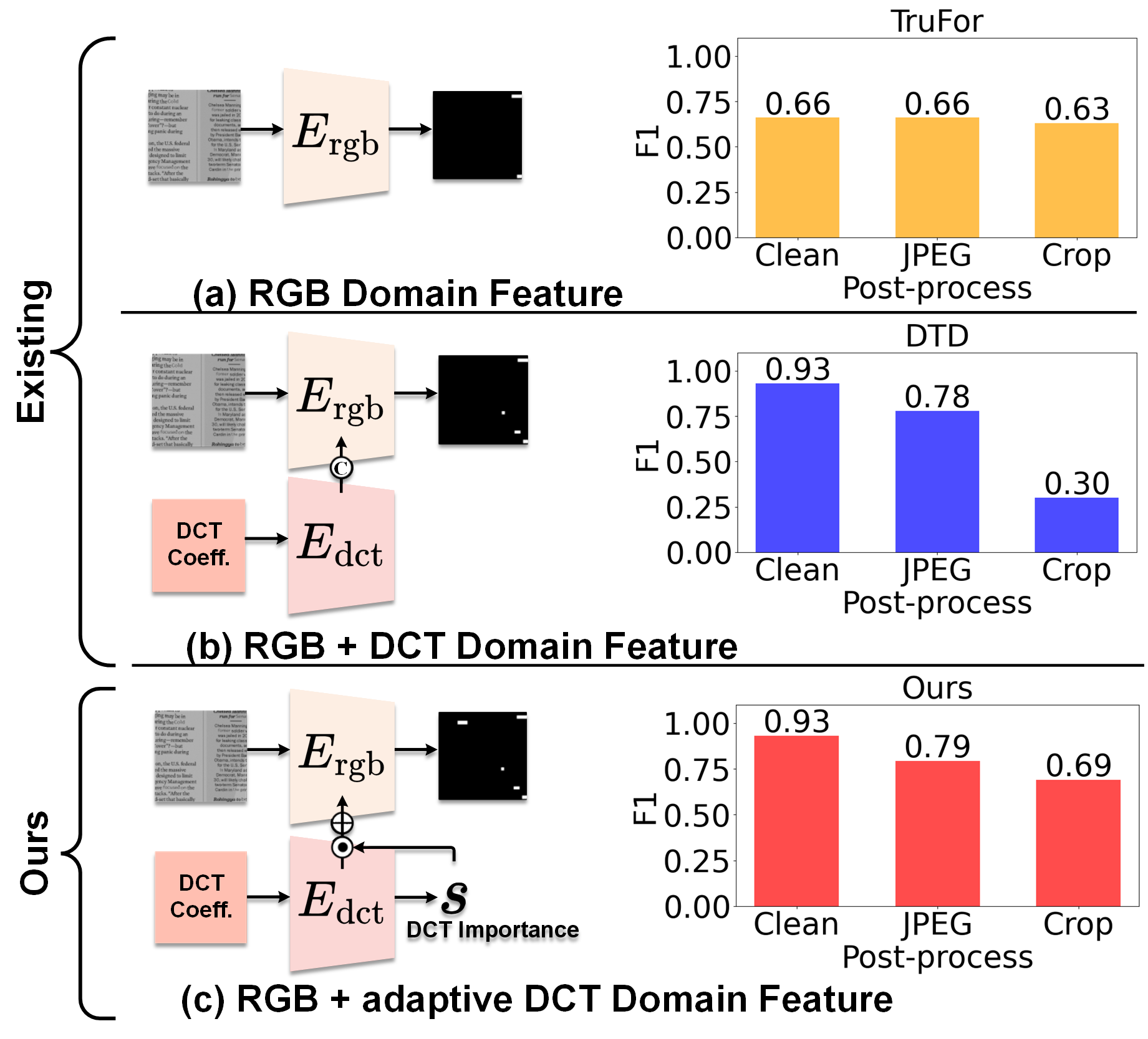}
    \vspace{-5pt}
    \caption{Comparison of ADCD-Net with existing methods: (a) Sole reliance on RGB forensic traces is insufficient for forgery detection. (b) Combining RGB and DCT features enhances effectiveness and robustness against JPEG compression but becomes severely vulnerable to disruptions in DCT block alignment. (c) ADCD-Net adaptively leverages DCT traces through a predicted score, simultaneously achieving desirable effectiveness and robustness against various distortions.}
    \label{fig:intro}
    \vspace{-8pt}
\end{figure}

\begin{figure}[t]
    \centering
    \includegraphics[width=0.48\textwidth]{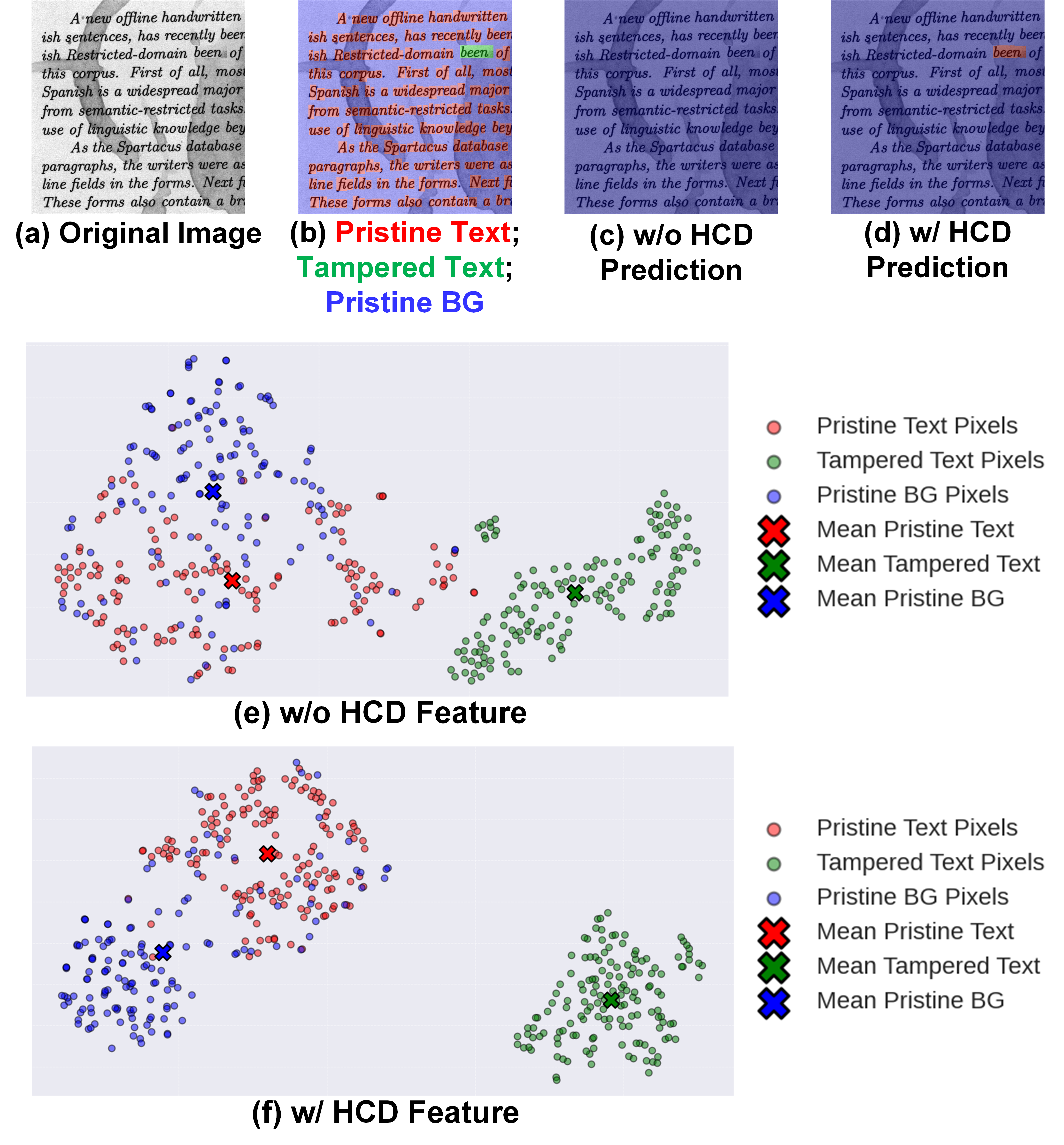}
    \vspace{-15pt}
    \caption{
    The strong intensity contrast between text and background (BG) (a \& b) causes pristine text pixels (red) to be undesirably drawn toward tampered text (green) in the feature space (c \& e) instead of aligning with the pristine BG (blue) ideally. The proposed Hierarchical Content Decoupling (HCD) mitigates this text-BG bias (d \& f). }
    \label{fig:hcd_intro}
    \vspace{-8pt}
\end{figure}

Forgery detection in natural images has been extensively studied, with various methods focusing on specific types of image forgery such as splicing~\cite{bi2019rru}, copy-move~\cite{li2018fast}, and inpainting~\cite{wu2021iid, li2023transformer}. Some approaches target particular forgery traces, including JPEG artifacts~\cite{lin2009fast, nikoukhah2019jpeg, bi2023self}, camera traces~\cite{cozzolino2019noiseprint, zheng2023exif}, and editing traces~\cite{guillaro2023trufor}, while others focus on learning general forgery features directly on RGB domain~\cite{wu2023rethinking, qu2024towards}. More recent solutions address complex and mixed types of forgery, including those involving transmission degradation and various post-processing operations, by detecting general or mixture of forgery artifacts~\cite{wu2022robust, dong2022mvss, kwon2022learning, liu2022pscc, guillaro2023trufor, zhang2024new}. However, forensic cues in natural images differ significantly from those in document images. Specifically, document images often feature uniform background (BG) and structured text with sharp contours and consistent textures. Tampered regions can be extremely small, blending seamlessly with their surroundings and thus making forgeries easy to execute but difficult to detect. These unique characteristics pose great challenges to accurate and robust forgery localization for documents.  

Substantial efforts have been made to localize forged areas in document images~\cite{taburet2022document, joren2022learning, shao2023progressive, qu2023towards, dong2024robust, zhang2023attention}. Current methods leverage various traces, such as JPEG artifacts~\cite{taburet2022document}, noise and texture traces~\cite{shao2023progressive}, or directly derive from the RGB domain~\cite{joren2022learning, dong2024robust}. Particularly,~\cite{qu2023towards} extracts forensic features from both RGB and DCT domains, achieving competitive detection performance on document images and exhibiting robustness against JPEG recompression. However, real-world document images often undergo various degradations, including cropping and resizing, which could obscure forensic features and even break JPEG grids alignment, making the existing DCT-based methods severely vulnerable in practice~\cite{bi2023self, guillaro2023trufor}. Additionally, existing methods do not fully utilize the distinctive characteristics of document images, further leading to inferior performance.  


To address the aforementioned challenges, we propose a novel forgery localization model, ADCD-Net, specifically designed for tampered document images. The key innovation of ADCD-Net lies in \textbf{\textit{adaptively}} utilizing forensic traces from both the RGB and DCT domains to enhance detection performance and robustness against various distortions; not limited to JPEG compression. This innovation is inspired by the fact that, while the \(8 \times 8\) block DCT-based features have proven effective over other form of frequency features (such as LoG~\cite{wang2022detecting}, Bayar~\cite{bayar2018constrained} and SRM~\cite{fridrich2012rich}) in prior works~\cite{kwon2022learning, qu2023towards, zhang2024new}, their reliability significantly decreases when the \(8 \times 8\) DCT block alignment is disrupted by operations like resizing or cropping~\cite{bi2023self, guillaro2023trufor}. Therefore, instead of completely discarding DCT features or fully relying on static DCT features (Fig.~\ref{fig:intro}~(a-b)), we propose adaptively considering DCT features as shown in Fig.~\ref{fig:intro}~(c). Specifically, adaptive DCT features are extracted through an optimized scoring mechanism that dynamically modulates its contributions, thereby providing robust input features for subsequent forgery decisions.



However, the multi-view features still suffer from significant text-BG bias as illustrated in Fig.~\ref{fig:hcd_intro}, which would lead to inferior localization accuracy. To resolve this challenge, we propose two key modules, Hierarchical Content Decoupling (HCD), and Pristine Prototype Estimation (PPE) that utilize the domain knowledge of the document images toward more accurate and robust forgery localization.
Specifically, the text-BG bias creates feature disparities that could obscure subtle tampering cues. The proposed HCD effectively separates forgery from content features across scales, reducing such text-BG bias and improving detection accuracy. 
Additionally, it is a reasonable assumption that most BG regions are pristine, as forgery often occurs in informative text regions. This phenomenon can be easily verified in many datasets, \eg \cite{qu2023towards, rossler2019faceforensics++}. The proposed PPE generates a prototype capturing pristine noise patterns. By comparing the pristine prototype with the entire image, we can effectively reveal subtle tampered areas, and enhance model performance and robustness. Our key contributions are summarized as follows:
\begin{itemize}
    \item We propose ADCD-Net, a robust document forgery localization model that adaptively fuses RGB and DCT traces, adjusting DCT feature contributions with a predicted alignment score. This enhances detection accuracy and robustness by mitigating the impact of alignment-disrupting operations such as resizing and cropping.
    \item We introduce HCD and PPE modules tailored to isolate forgery features from document content and estimate pristine noise patterns to uncover subtle tampered areas, further improving detection performance and robustness.
    \item Extensive experiments show that our ADCD-Net surpasses state-of-the-art (SOTA) methods by an average gain of 20.79\% across various distortion types.
\end{itemize}


\section{Related Works}

\subsection{Forgery Localization on Natural Images}

Extensive methods for image manipulation detection focus on specific types of forgery, such as splicing~\cite{bi2019rru}, copy-move~\cite{li2018fast}, and inpainting~\cite{wu2021iid, li2023transformer}, or targeted forensic traces such as JPEG artifacts~\cite{lin2009fast, nikoukhah2019jpeg, bi2023self}, camera traces~\cite{cozzolino2019noiseprint, zheng2023exif}, and editing traces~\cite{guillaro2023trufor}. More robust approaches target complex and mixed forgeries. For example, CAT-Net~\cite{kwon2022learning} introduces a learnable module to extract JPEG double quantization artifacts from the DCT coefficients and combines features from both RGB and DCT domains to localize tampered regions. MVSS-Net~\cite{dong2022mvss} exploits boundary artifacts and noise with multi-scale supervision. OSN~\cite{wu2022robust} introduces a robust training scheme based on estimated and adversarial noise to handle challenges posed by OSN lossy operations. Trufor~\cite{guillaro2023trufor} learns a noise-sensitive fingerprint in a self-supervised manner and fuse the fingerprint with the RGB domain features to predict forgery mask. Other approaches directly target forgery feature learning in the RGB domain. PSCC-Net~\cite{liu2022pscc} employs a progressive framework to predict forged masks at multiple scales. FOCAL~\cite{wu2023rethinking} introduces a novel perspective where forged/pristine pixels are treated as relative concepts within an image, and propose a within-image contrastive loss and clustering inference for robust detection. APSC-Net~\cite{qu2024towards} constructs a large-scale forgery image dataset and incorporates the adaptive, self-calibration modules into the decoder for improved localization accuracy. Nevertheless, these methods often struggle to detect tampering in document images, as the tampered regions can be seamlessly blended into the uniform document BGs and structured texts. 

\subsection{Forgery Localization on Document Images}
Traditional methods for document tampering detection typically target simple, clean documents. For example, ~\cite{bertrand2015conditional} uses font features to distinguish tampered texts, while ~\cite{van2013text} analyzes text line alignment to locate forgery areas. Recent learning-based approaches have improved performance on photographed documents. For instance, ~\cite{taburet2022document} employs a CNN to detect double compression for forgery localization, and ~\cite{joren2022learning} uses a graph attention mechanism for tampered text detection. PS-Net~\cite{shao2023progressive} combines noise, texture, and RGB traces with progressive supervision to better identify tampered regions, and MA-Net~\cite{dong2024robust} adopts a two-stage model that first enhances forgery traces via reconstruction and then localizes tampered areas with a multi-scale attention network. However, methods that rely solely on RGB traces often fall short in practical scenarios. In contrast, DTD~\cite{qu2023towards} constructs a tampered document dataset and fuses RGB with DCT features for recompressed images. Yet, DTD reliance on DCT makes it vulnerable to disruptions in \(8\times8\) block alignment, and overall, existing methods struggle with challenges like recompression and resizing that obscure forensic evidence.

\section{Method}

\begin{figure*}[h]
    \centering
    \includegraphics[width=0.98\textwidth]{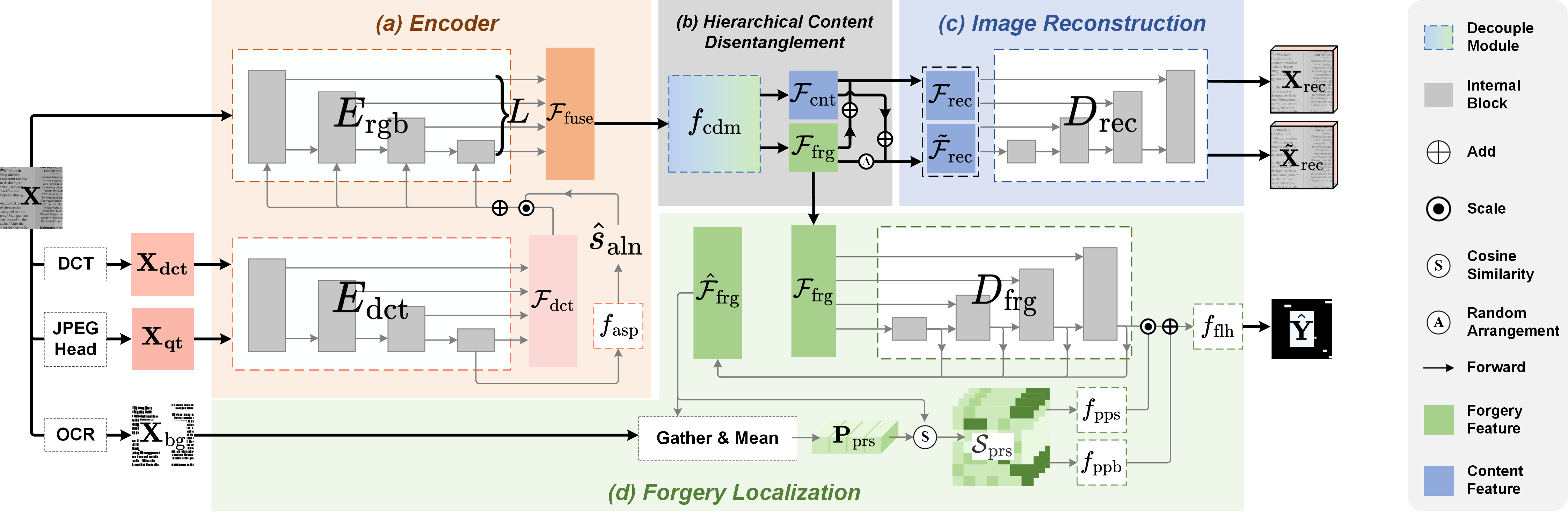}
    \vspace{-5pt}
    \caption{Overview of ACDC-Net, consisting of (a) adaptive RGB-DCT encoder \(E\) extracting multi-scale features from RGB and DCT domains; (b-c) HCD module \(f_\text{cdm}\) disentangling forgery and content features and the reconstruction branch reconstructs original image based on content features; (d) localization branch locates the tampered region based on decoupled forgery features.}
    \label{fig:model_overview}
    \vspace{-5pt}
\end{figure*}

We propose ADCD-Net, a specialized forgery localization model designed to address the unique challenges posed by the inherent characteristics of document images. Existing DCT-based methods fuse the DCT features without considering their importance, making them vulnerable to alignment disruptions. In contrast, we \textbf{\textit{selectively}} incorporate the DCT feature using a learnable score, which essentially measures its appropriateness and hence enhances robustness. Nevertheless, the fused RGB-DCT feature in this way still could be affected by strong text-BG bias, leading to inferior localization performance. To mitigate such a bias, we introduce a HCD module that reduces text-BG interference. Additionally, we leverage critical information from the document's BG through a PPE module. The estimated pristine prototype captures the noise pattern of untampered areas, which could be ultimately used to further boost the localization performance.     


The overall of ADCD-Net, depicted in \cref{fig:model_overview}, adopts a U-shaped network with dual decoders for content decoupling and forgery localization. Given an input JPEG image \(\mathbf{X} \in \mathbb{R}^{H \times W \times 3}\), we perform the \(8 \times 8\) block-based DCT transformation on the Y-channel of \(\mathbf{X}\) to obtain the quantized DCT coefficients \(\mathbf{X}_{\text{dct}} = \mathcal{DCT}(\mathbf{X}_Y) \in \mathbb{N}^{H \times W}\) and extract the quantization table \(\mathbf{X}_{\text{qt}} \in \mathbb{N}^{H \times W}\) from the JPEG header. The encoder \(E\) processes the inputs to capture adaptive, multi-scale forgery cues from both RGB and DCT domains, yielding \(\mathcal{F}_{\text{fuse}} = E(\mathbf{X}, \mathbf{X}_{\text{dct}}, \mathbf{X}_{\text{qt}})\). Specifically, we obtain the fuse feature at level-\(i\) by \(\mathbf{F}^{i}_{\text{rgb}} + \hat{s}_{\text{aln}} \cdot \mathbf{F}^{i} _{\text{dct}}\), in which \(\mathbf{F}_{\text{rgb}}\), \(\mathbf{F}_{\text{dct}}\), and \(\hat{s}_{\text{aln}}\) are the RGB feature, DCT feature, and the learned DCT weight. \(\mathcal{F}_{\text{fuse}}\) is then fed to the HCD module \(f_{\text{cdm}}\), which disentangles \(\mathcal{F}_{\text{fuse}}\) into content and forgery features \([\mathcal{F}_{\text{cnt}}, \mathcal{F}_{\text{frg}}] = f_{\text{cdm}}(\mathcal{F}_{\text{fuse}})\). The reconstruction inputs \(\mathcal{F}_{\text{rec}}\) and \(\tilde{\mathcal{F}}_{\text{rec}}\), derived from \(\mathcal{F}_{\text{cnt}}\) and \(\mathcal{F}_{\text{frg}}\), are processed by the reconstruction decoder \(D_{\text{rec}}\) to recover the original image and DCT coefficients \(\mathbf{X}_{\text{rec}}\) and  \(\tilde{ \mathbf{X}}_{\text{rec}}\). Simultaneously, the forgery features \(\mathcal{F}_{\text{frg}}\) are sent to the localization decoder \(D_{\text{frg}}\) to generate the forgery localization map \(\hat{\mathbf{Y}} \in \{0,1\}^{H \times W}\). During the inference phase, \(D_{\text{rec}}\) is discarded, leaving the remaining networks for inference.

\subsection{Adaptive RGB-DCT Encoder}
As aforementioned, DCT forensic trace is crucial for detection accuracy and robustness against recompression~\cite{kwon2022learning, qu2023towards}; but fragile to various post-processing that disrupts the \(8\times8\) block alignment~\cite{bi2023self, guillaro2023trufor}, such as cropping, resizing. Fig.~\ref{fig:model_overview}~(a) illustrates our adaptive RGB-DCT encoder \(E\), which leverages DCT domain features while mitigating their sensitivity. The encoder consists of two subnetworks: one extracts RGB features \(E_{\text{rgb}}(\mathbf{X})\) to produce the fused feature set \(\mathcal{F}_{\text{fuse}} = \{\mathbf{F}^{i}_{\text{fuse}}\}_{i=1}^{L}\), and the other extracts DCT features \(E_{\text{dct}}(\mathbf{X}_{\text{dct}}, \mathbf{X}_{\text{qt}})\) to yield \(\mathcal{F}_{\text{dct}} = \{\mathbf{F}^{i}_{\text{dct}}\}_{i=1}^{L}\). For each level \(i\), the DCT feature \(\mathbf{F}^{i}_{\text{dct}}\) is adaptively fused into the corresponding RGB feature using a predicted alignment score \(\hat{s}_{\text{aln}} = f_{\text{asp}}(\mathbf{F}^{L}_{\text{dct}}) \in (0,1)\), which controls its contribution. Specifically, the fusion is defined as

\begin{equation}
\begin{split}
  \mathbf{F}^{i+1}_{\text{fuse}} = f^{i}_{\text{fuse}}\Big(\mathbf{F}^{i}_{\text{rgb}} + f_{\text{asp}}(\mathbf{F}^{L}_{\text{dct}}) \cdot \mathbf{F}^{i}_{\text{dct}}\Big),
\end{split}
\label{eq:important}
\end{equation}

where \(f^{i}_{\text{fuse}}\) denotes the level-\(i\) block of \(E_{\text{rgb}}\) and \(L\) is the total number of blocks. The alignment score \(\hat{s}_{\text{aln}}\) is predicted by a head appended to the last layer of \(E_{\text{dct}}\) and is applied uniformly across all scales. We cast its estimation as a classification task, with the output probability serving as \(\hat{s}_{\text{aln}}\). Given forgery images with well-aligned DCT blocks, we disrupt the alignment using augmentations like random resizing, cropping, and pixel shifting\footnote{We crop and shift pixels \(n \bmod 8 \neq 0\) to avoid block-aligned output.} to create non-aligned samples. The ground-truth label \(s_{\text{aln}}\) can be readily assigned, with non-aligned samples labeled as ``0", while aligned ones as ``1". 

Remark: Our adaptive DCT encoder builds on the MoE principle \cite{jacobs1991adaptive, shazeer2017outrageously}, showing its potential to generalize across various forensic traces beyond just DCT features. In a typical MoE setup, specialized expert subnetworks are weighted by a gating mechanism. Here, \(E_{\text{dct}}\) targets compression artifacts while \(E_{\text{rgb}}\) captures robust and general forensic cues, with an adaptive routing function \(f_{\text{asp}}(\mathbf{F}^{L}_{\text{dct}})\) controlling the DCT contribution for each sample. Unlike existing MoE methods that learn routing weights using the prediction-GT \cite{jacobs1991adaptive} or sparsity \cite{shazeer2017outrageously} objectives, we optimize \(f_{\text{asp}}\) with an explicit classification task to prioritize DCT alignment. Although both DTD~\cite{qu2023towards} and our ADCD-Net leverage RGB and DCT traces, DTD overfits to DCT features and is vulnerable to post-processing (see Fig.~\ref{fig:non_align} DTD). Even with RGB input, DTD is significantly less robust than other RGB-based detectors such as TruFor and ADCD-Net. This indicates that DTD's simple RGB-DCT concatenation fails to prioritize critical features. In contrast, our adaptive DCT encoder with the MoE structure enables the model to rely primarily on RGB cues when DCT traces are weak, and vice versa (see Fig.~\ref{fig:agm}).



\subsection{Hierarchical Content Decoupling}
To mitigate our observed text-BG bias (see Fig.~\ref{fig:hcd_intro}), we propose the HCD module \(f_{\text{cdm}}\) to disentangle content and forgery features at multiple scales. This disentanglement is possible because forgery cues stem from subtle inconsistencies or noise rather than the intrinsic image content, while content features are closely tied to the visual structure (see Fig.~\ref{fig:hcd_vis}). As depicted in \cref{fig:model_overview} (b), after we obtain \(\mathcal{F}_{\text{fuse}}\), the decoupling module \(f_{\text{cdm}}\) is employed to map \(\mathbf{F}^{i}_{\text{fuse}}\) into content \(\mathbf{F}^{i}_{\text{cnt}}\) and forgery feature \(\mathbf{F}^{i}_{\text{frg}}\) at each \(i\) scale, such that:
\begin{equation}
\begin{split}
  f_{\text{cdm}}(\mathbf{F}^{i}_{\text{fuse}}) = [\mathbf{F}^{i}_{\text{cnt}}, \mathbf{F}^{i}_{\text{frg}}] \in \mathbb{R}^{H \times W \times 2C}. 
\end{split}
\label{eq:important}
\end{equation}
For efficiency, \(f_{\text{cdm}}\) at each scale is implemented as a multi-layer perceptron (MLP). The decoupled process is performed at each scale of the input features, resulting in \([\mathcal{F}_{\text{cnt}}, \mathcal{F}_{\text{frg}}] = f_{\text{cdm}}(\mathcal{F}_{\text{fuse}})\). Then, \(\mathcal{F}_{\text{cnt}}\)  and \(\mathcal{F}_{\text{frg}}\) are passed to the following reconstruction and localization decoders. It is important to note that the decoupling process is not solely carried out by \(f_{\text{cdm}}\); instead, it is achieved through the following reconstruction and localization decoders and the training objectives.  Consequently, a lightweight MLP is sufficient within the decoupling module.  

As shown in \cref{fig:model_overview} (c), the decomposed features \(\mathcal{F}_{\text{cnt}}\) and \(\mathcal{F}_{\text{frg}}\) are used to reconstruct the original image \(\mathbf{X}\) and DCT coefficients \(\mathbf{X}_{\text{dct}}\). To reduce the content influence in the forgery features \(\mathcal{F}_{\text{frg}}\), we ensure that the content information is primarily captured by \(\mathcal{F}_{\text{cnt}}\). This is achieved by randomly shuffling the spatial arrangement of \(\mathcal{F}_{\text{frg}}\) to form \(\tilde{\mathcal{F}}_{\text{frg}}\) before fusing it back with \(\mathcal{F}_{\text{cnt}}\), and then requiring the network to reconstruct \([\mathbf{X}, \mathbf{X}_{\text{dct}}]\). Specifically, \(\mathcal{F}_{\text{rec}} = \{\mathbf{F}^{i}_{\text{cnt}} + \mathbf{F}^{i}_{\text{frg}}\}^{L}_{i=1}\) and \(\tilde{\mathcal{F}}_{\text{rec}} = \{\mathbf{F}^{i}_{\text{cnt}} + \tilde{\mathbf{F}}^{i}_{\text{frg}}\}^{L}_{i=1}\) serve as two distinct inputs to the reconstruction decoder \(D_{\text{rec}}\) for generating \(\mathbf{X}_{\text{rec}} = D_{\text{rec}}(\mathcal{F}_{\text{rec}})\) and \(\tilde{\mathbf{X}}_{\text{rec}} = D_{\text{rec}}(\tilde{\mathcal{F}}_{\text{rec}})\), which are respectively the reconstructed versions of \(\mathbf{X}\) and \(\mathbf{X}_{\text{dct}}\). 


Remark: Existing content disentanglement methods~\cite{liang2022exploring, yan2023ucf, ye2024decoupling, li2022artifacts, liu2020disentangling} are mainly designed for face forgery detection. Motivated by our observed text-BG bias in Fig.~\ref{fig:hcd_intro}~(e), we extend content disentanglement to dense prediction tasks with multi-modal inputs by proposing the HCD module. This module hierarchically decouples content, preventing data leakage through shortcuts in the U-shaped network and effectively disentangling multi-scale features to accommodate varying text scales in documents.


\subsection{Forgery Localization}

We introduce the PPE module, which injects prior knowledge of untampered regions to enhance performance. PPE can be selectively employed when high-confidence pristine areas exist (\eg the BG of deepfake portraits and forged documents, which is less informative and predominantly pristine). As shown in Fig.~\ref{fig:model_overview}~(d), an OCR model \(f_{\text{ocr}}\) is used to extract the BG of the image \(\mathbf{X}_{\text{bg}}=f_{\text{ocr}}(\mathbf{X}) \in \{0, 1\}^{H \times W}\), in which “0" is the text area while “1" is the BG. The estimated pristine prototype on level-\(i\)  feature is computed by 
\begin{equation}
  \mathbf{p}^{i}_{\text{prs}} = \frac{\sum_{h,w} \mathbf{X}_{\text{bg}}(h,w) \hat{\mathbf{F}}^{i}_{\text{frg}}(h,w)}{\sum_{h,w} \mathbf{X}_{\text{bg}}(h,w)},
  \label{eq:important}
\end{equation}
in which \(h\) and \(w\) are the index of height and width, and \(\hat{\mathbf{F}}^{i}_{\text{frg}}\) is the output of the level-\(i\) block in \(D_{\text{frg}}\). Then, the estimated pristine map can be obtained by
\begin{equation}
  \mathbf{S}^{i}_{\text{prs}}(h,w) = \frac{\hat{\mathbf{F}}^{i}_{\text{frg}}(h,w) \cdot \mathbf{p}^{i}_{\text{prs}}}{\|\hat{\mathbf{F}}^{i}_{\text{frg}}(h,w)\| \|\mathbf{p}^{i}_{\text{prs}}\|}.
  \label{eq:important}
\end{equation}

As can be observed in Fig.~\ref{fig:hcd_intro}~(f), by incorporating the HCD, the pristine prototype (in blue cross) is more accurately located in the pristine cluster. The pristine map is computed at multiple scales, resulting in \(\mathcal{S}_{\text{prs}} = \{\mathbf{S}^{i}_{\text{prs}}\}^{L}_{i=1}\). The maps \(\mathcal{S}_{\text{prs}}\) modulate the penultimate feature via element-wise scaling and biasing. Two MLP layers, \(f_\text{pps}\) and \(f_\text{ppb}\), convert \(\mathcal{S}_{\text{prs}}\) into a scale and a bias, respectively. This yields
\begin{equation}
    \hat{\mathbf{Y}} = f_{\text{flh}}\Big( \hat{\mathbf{F}}^{L}_{\text{frg}} \cdot f_\text{pps}(\mathcal{S}_{\text{prs}}) + f_\text{ppb}(\mathcal{S}_{\text{prs}})\Big),
\label{eq:important}
\end{equation}
where \(\hat{\mathbf{F}}^{L}_{\text{frg}}\) is the penultimate feature and \(f_{\text{flh}}\) is the segmentation head producing the predicted tampered map \(\hat{\mathbf{Y}}\).

\subsection{Training Objectives}

\begin{figure*}
    \centering
    \includegraphics[width=0.95\textwidth]{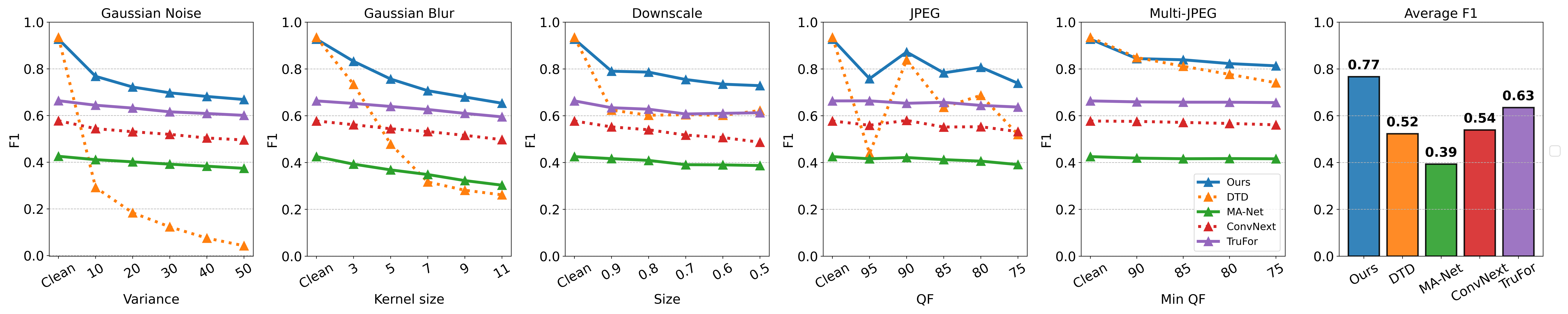}
    \vspace{-5pt}
    \caption{(Axes 1-5) The localization performance in F1 score of different competitors after undergoing various distortions like Gaussian noise, Gaussian blur, downscale, JPEG compression, and multiple JPEG compression. (Axis 6) The average F1 score across all distortions.}
    \label{fig:robust}
\end{figure*}

\begin{table*}[h]
\small
\begin{center}
\renewcommand{\arraystretch}{0.8} 
\begin{tabular}{l|ccc|ccc|ccc|c|c}
    \hline

    \multirow{2}{*}{Method} & \multicolumn{3}{c|}{\texttt{Test} (30,000)} & \multicolumn{3}{c|}{\texttt{FCD} (2,000)} & \multicolumn{3}{c|}{\texttt{SCD} (18,000)} & \multirow{2}{*}{AVG F1} & \multirow{2}{*}{Params (M)}  \\ 

    \cline{2-10}
    
    & P & R & F1 & P & R & F1 & P & R & F1 & \\

    \hline
    
    MVSS-Net~\cite{dong2022mvss} & .494 & .383 & .431 & .480 & .381 & .424 & .478 & .366 & .414 & .423 & 143 \\

    CAT-Net~\cite{kwon2022learning} & .768 & .680 & .721 & .795 & .695 & .741 & .674 & .665 & .670 & .702 & 114 \\

    \(\text{TruFor}^{*}\)~\cite{guillaro2023trufor} & .176 &.160 & .168 &.230 &.072 &.109 &.257 &.311 &.281 &.205 & 67 \\

    TruFor~\cite{guillaro2023trufor} & .673 & .710 & .691 & .659 & .675 & .667 & .528 & .678 & .594 & .655 & 67 \\

    \(\text{APSC-Net}^{*}\)~\cite{qu2024towards} & .137 & .182 & .156 & .259 & .272 & .266 & .258 & .343 & .295 & .211 & 143 \\

    \hline
    
    ConvNext-Uper~\cite{liu2022convnet} & .676 & .591 & .630 & .606 & .450 & .517 & .550 & .521 & .535 & .589 & 123 \\
    
    MA-Net~\cite{dong2024robust} & .459 & .473 & .466 & .335 & .405 & .367 & .349 & .508 & .414 & .445 & \underline{24} \\

    DTD~\cite{qu2023towards} & \textbf{.814} & .771 & .792 & \underline{.849} & \underline{.786} & \underline{.816} & \textbf{.745} & .762 & \textbf{.754} & .778 & 66 \\

    PS-Net~\cite{shao2023progressive} & .698 & \textbf{.972} & \underline{.798} & .716 & \textbf{.987} & \textbf{.827} & .659 & \textbf{.909} & \underline{.741} & \underline{.785} & - \\
    
    ACDC-Net (Ours) & \underline{.789} & \underline{.823} & \textbf{.806} & \textbf{.866} & .770 & .815 & \underline{.690} & \underline{.799} & .740 & \textbf{.787} & \textbf{23} \\

    \hline
\end{tabular}
\vspace{-5pt}
\caption{Comparison on localization performance across 3 test sets, with recompressed samples using the quality factors from DocTamper. ``P'', ``R'', and ``F1'' represent precision, recall, and F1 score. Models marked with ``\(*\)'' used the checkpoint from the original repository. The best result is in \textbf{bold}, and the second-placed is \underline{underlined}. Since PS-Net is not accessible, the number of parameters is excluded.}
\label{table:multu_jpeg}
\end{center}
\vspace{-20pt}
\end{table*}

ADCD-Net is trained in an end-to-end fashion by using the following loss function:
\begin{equation}
  \mathcal{L} = \lambda_{\text{aln}}\mathcal{L}_{\text{aln}} + \lambda_{\text{rec}}\mathcal{L}_{\text{rec}} +  \lambda_{\text{frg}}\mathcal{L}_{\text{frg}} + \lambda_{\text{con}}\mathcal{L}_{\text{con}}.
\label{eq:important}
\end{equation} Here, the alignment score loss \(\mathcal{L}_{\text{aln}}\) ensures the accuracy of the predicted alignment score \(\hat{s}_{\text{aln}}\). The image reconstruction loss \(\mathcal{L}_{\text{rec}}\) maintains the quality of the reconstructed image, implicitly validating the feature disentanglement. The forgery localization loss \(\mathcal{L}_{\text{frg}}\) ensures accurate prediction of the tampered mask. Lastly, the within-image contrastive loss \(\mathcal{L}_{\text{con}}\) amplifies the distinction between pristine and forged pixels in the feature domain. The details on the calculation of these four loss terms are given below.

\noindent\textbf{Alignment score loss} \(\mathcal{L}_{\text{aln}}\). The cross-entropy loss \(\mathcal{L}_\text{aln}\) is computed between the prediction score \(\hat{s}_{\text{aln}}\) and the ground-truth \(s_{\text{aln}}\), to accurately and dynamically control the magnitude of the DCT features toward more robust localization.

\noindent\textbf{Image reconstruction loss} \(\mathcal{L}_{\text{rec}}\). The \(\ell_1\) loss is used to ensure the reconstructed content is close to the original:
\begin{equation}
\begin{split}
  \mathcal{L}_\text{rec} =  \| [\mathbf{X}, \mathbf{X}_{\text{dct}}] - D_{\text{rec}}(\{\mathbf{F}^{i}_{\text{cnt}} + 
  \mathbf{F}^{i}_{\text{frg}}\}^{L}_{i=1}) \| + \\ 
  \| [\mathbf{X}, \mathbf{X}_{\text{dct}}] - D_{\text{rec}}(\{\mathbf{F}^{i}_{\text{cnt}} + \tilde{\mathbf{F}}^{i}_{\text{frg}}\}^{L}_{i=1}) \|.
\end{split}
\label{eq:important}
\end{equation}
\noindent\textbf{Forgery localization loss} \(\mathcal{L}_{\text{frg}}\). The loss \(\mathcal{L}_{\text{frg}}\) is used to compute the error between the prediction \(\hat{\mathbf{Y}}\) and the ground-truth forgery mask \(\mathbf{Y}\) with the cross-entropy loss and Lovase loss~\cite{berman2018lovasz}, in which \(\mathcal{L}_\text{frg} = \lambda_{\text{ce}}\mathcal{L}_\text{ce}(\hat{\mathbf{Y}}, \mathbf{Y}) + \mathcal{L}_\text{lov}(\hat{\mathbf{Y}}, \mathbf{Y})\).

\noindent\textbf{FOCAL loss} \(\mathcal{L}_{\text{con}}\). Inspired by FOCAL~\cite{wu2023rethinking}, with the idea that forged/pristine pixels are relative concepts within an image, we adopt the within-image contrastive loss to further enlarge the discrepancy between pristine and forged pixels. The contrastive loss is computed on multi-scale forgery features \(\hat{\mathcal{F}}_{\text{frg}} = \{\hat{\mathbf{F}}^{i}_{\text{frg}}\}^{L}_{i=1}\). Given the extreme imbalance between pristine and forged areas, we use the SupCon loss~\cite{khosla2020supervised, zhu2022balanced} to balance the influence of each class. Specifically, the contrastive loss for the \(i\)-th level feature \(\hat{\mathbf{F}}^{i}_{\text{frg}}\) is formulated as:
\begin{equation}
    \mathcal{L}^{b}_{\text{con},i} = \sum_{\mathbf{z} \in \mathcal{Z}} \frac{-1}{|\mathcal{P}|} \sum_{\mathbf{p} \in \mathcal{P}} \log \frac{\exp ( \mathbf{z} \cdot \mathbf{p})} {\sum_{j\in\mathcal{Y}} \sum_{\mathbf{a} \in \mathcal{A}_j} \exp ( \mathbf{z} \cdot \mathbf{a})}.
\label{eq: SD loss}
\end{equation}
\noindent where \(\mathcal{Z}\) denotes the entire pixel set for \(\hat{\mathbf{F}}^{i}_{\text{frg}}\). The positive set \(\mathcal{P}\) for a pixel \(\mathbf{z}\) is defined as \( \mathcal{P} = \{ \mathbf{p} \in \mathcal{Z} \mid y_{\mathbf{p}} = y_{\mathbf{z}} \} \setminus \{ \mathbf{z}\}\). The label set \(\mathcal{Y} = \{0, 1\}\) represents pristine and tampered labels and the anchor set of \(\mathbf{z}\) for label \(y_j\) is defined as \(\mathcal{A}_j = \{ \mathbf{a} \in \mathcal{Z} \mid y_{\mathbf{a}} = y_j \} \setminus \{ \mathbf{z}\}\). Since \(|\mathcal{Z}|\) is typically large, we sample \(\min(n, |\mathcal{A}_j|+1)\) pixels for \(y_j\) class in \(\mathcal{Z}\) for efficiency. The multi-scale contrastive loss \(\mathcal{L}_{\text{con}}\) is then obtained by summing over all scales and batch samples: \(\mathcal{L}_{\text{con}} = \sum^{L}_{i=1} \sum_{b\in \mathcal{B}} \mathcal{L}^{b}_{\text{con}, i}\).

\section{Experiments}
\subsection{Implementation Details}
ADCD-Net uses the Restormer encoder~\cite{zamir2022restormer} as \(E_{\text{rgb}}\) and the Restormer decoder as \(D_{\text{rec}}\) and \(D_{\text{frg}}\), initialized with DocRes~\cite{zhang2024docres}. For \(E_{\text{dct}}\), we adopt FPH from DTD~\cite{qu2023towards}, and we use CRAFT~\cite{baek2019character} as the OCR model. Our model is trained on the DocTamper~\cite{qu2023towards} training set and evaluated on the cross-domain test sets (\texttt{Test}, \texttt{FCD}, and \texttt{SCD}). More details are presented in the supplementary.

\subsection{Competitors}
We evaluate our method against general forgery localization models: MVSS-Net~\cite{dong2022mvss}, CAT-Net~\cite{kwon2022learning}, TruFor~\cite{guillaro2023trufor}, and APSC-Net~\cite{qu2024towards}, as well as document-specific models: MA-Net~\cite{dong2024robust}, DTD~\cite{qu2023towards}, and PS-Net~\cite{shao2023progressive}. We also implement ConvNext-Uper~\cite{liu2022convnet, xiao2018unified}, the 1st place of the 2022 Real-World Image Forgery Localization Challenge\footnote{https://github.com/Junjue-Wang/Rank1-Ali-Tianchi-Real-World-Image-Forgery-Localization-Challenge}\footnote{https://tianchi.aliyun.com/competition/entrance/531945/introductio}.

\subsection{Accuracy and Robustness Evaluation}

\begin{figure}[]
    \centering
    \includegraphics[width=0.45\textwidth]{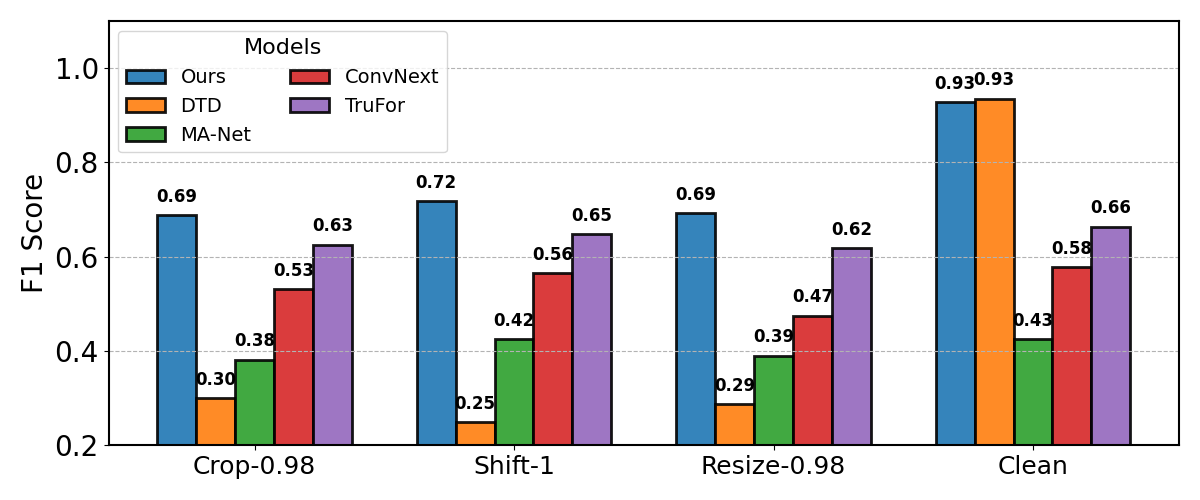}
    \vspace{-8pt}
    \caption{Localization performance (F1 score) across various DCT block misalignment perturbations.}
    \label{fig:non_align}
    \vspace{-15pt}
\end{figure}

We first evaluate the robustness of the forgery localization models. Fig.~\ref{fig:robust} shows the model performance average on the three test sets subjected to five distortions, with severity increasing from left to right (Clean indicates no distortion). The multi-JPEG distortion randomly compresses images 1 to 3 times at a minimum quality factor~\cite{qu2023towards}. Our method outperforms the second-best, TruFor, by an average of \textbf{\textit{20.79\% across all distortions}}. Although DTD performs well under multi-JPEG, its F1 score \textbf{\textit{drops by 95.50\% and 71.96\%}} at the highest noise and blur levels, respectively, due to the disrupted DCT statistics. In contrast, RGB-based models TruFor, ConvNext-Uper, and MA-Net exhibit strong robustness, though they have lower base accuracy.

We explore the DCT feature limitations by introducing slight structural disruptions that break the DCT block alignment. The disruptions include cropping 2\% off the image edges, shifting by 1 pixel, and scaling to 98\% of the original size. Fig.~\ref{fig:non_align} shows that while DTD achieves an F1 score of 0.935 on undistorted images, its performance \textbf{\textit{drops by 67.95\%, 73.34\%, and 69.28\%}} with these disruptions. In contrast, our method using adaptive DCT feature modulation shows only moderate declines of 25.74\%, 22.66\%, and 25.39\%. Although RGB-based methods like TruFor are robust to such changes, their overall performance is still lower than that of our ACDC-Net.

We also evaluate the localization performance under the DocTamper setting\footnote{https://github.com/qcf-568/DocTamper}, where all test images undergo multiple JPEG compressions with a minimum quality factor of 75 (see Table~\ref{table:multu_jpeg}). Our method outperforms the second-best PS-Net by a small margin (.002) while using the lowest number of parameters (23M). DCT-based methods like CAT-Net and DTD achieve average F1 scores of 0.702 and 0.778, respectively, surpassing most RGB-based methods such as TruFor and MA-Net. However, as demonstrated above, these DCT-based models are vulnerable to post-processing operations that disrupt DCT coefficient statistics, leading to significant performance drops.

\begin{table}[]
\small
\begin{center}
\renewcommand{\arraystretch}{0.8} 
\begin{tabular}{l|ccccc}
    \hline

    Metric & TurFor & MA-Net & DTD & ACDC-Net \\

    \hline
    F1 & .655 & .445 & \underline{.779} & \textbf{.808} \\
    FAR & \underline{.027}  & .029 & .058 & \textbf{.012} \\
    
    \hline
\end{tabular}
\vspace{-8pt}
\caption{F1 and false alarm rate (FAR) on pristine images.}
\label{table:fa}
\end{center}
\vspace{-10pt}
\end{table}  

\begin{table}[]
\begin{center}
\small
\renewcommand{\arraystretch}{0.8} 
\setlength{\tabcolsep}{1.3mm}{
\begin{tabular}{l|cccccccc}
    \hline

    \(\hat{s}_{\text{aln}}\) & N-30 & B-7 & D-.7 & J-85 & C-.98 & S-1 & R-98 \\

    \hline

    \(f_{\text{asp}}\) & \textbf{.697} & \textbf{.707} & \textbf{.755} & \textbf{.783} & \textbf{.689} & \textbf{.717} & \textbf{.692} \\
     
    \(0\) & \underline{.688} & \underline{.678} & .658 & .717 & \underline{.681} & \underline{.711} & \underline{.682} \\
    
    \(1\) & .671 & .572 & \underline{.724} & \underline{.764} & .621 & .687 & .632 \\
    
    \hline

    \(\Delta_{\text{1vs0}}\) & \textcolor{Red}{-.017} & \textcolor{Red}{-.105} & \textcolor{Green}{+.065} & \textcolor{Green}{+.047} & \textcolor{Red}{-.059} & \textcolor{Red}{-.024} & \textcolor{Red}{-.050} \\

    \hline
\end{tabular}
}
\vspace{-8pt}
\caption{Impact of the DCT modulation score \(\hat{s}_{\text{aln}}\) on the F1 score across the distortions. \(\Delta_{\text{1vs0}}\) represents the F1 gap between \(\hat{s}_{\text{aln}}=0\) and \(\hat{s}_{\text{aln}}=1\). N-30: Noise-30; B-7: Blur-7; D-.7: Down-0.7; J-85: JPEG-85; C-.98: Crop-0.98; S-1: Shift-1; R-98: Resize-0.98.}
\label{table:align_score}
\end{center}
\vspace{-20pt}
\end{table}

Evaluating false alarm rates (FAR) on pristine images is crucial. We tested 1,000 pristine document images from the 2023 Detecting Tampered Text in Images Tianchi Competition dataset~\cite{luo2023icdar}. As shown in Table~\ref{table:fa}, our model achieves the lowest FAR (1.2\%) while maintaining the highest F1.

The supplementary includes quantitative evaluations under more severe degradations and qualitative assessments.


\subsection{Analysis on Proposed Modules}

\textbf{Effect on the DCT modulation factor \(\hat{s}_{\text{aln}}\).} As shown in Table~\ref{table:align_score}, we evaluate performance by setting \(\hat{s}_{\text{aln}}\) to 0, 1, and its predicted value \(\hat{s}_{\text{aln}} = f_{\text{asp}}(\mathbf{F}^{L}_{\text{dct}})\) across various distortions. We report the performance difference \(\Delta_{\text{1vs0}}\) between \(\hat{s}_{\text{aln}}=1\) (fully using DCT features) and \(\hat{s}_{\text{aln}}=0\) (disabling DCT features), highlighting the impact under different conditions. Overall, fusing DCT features based on the predicted score consistently yields the highest F1 score. Notably, when \(\hat{s}_{\text{aln}}=0\), the model outperforms the \(\hat{s}_{\text{aln}}=1\) case in five distortion types, with an average F1 gain of 8.37\%. DCT features are less effective under block-misalignment distortions (e.g., cropping, pixel-shifting, resizing) and under blurring or noise addition, which disrupt both low- and high-frequency DCT statistics. In contrast, for distortions like downscaling and JPEG compression, where DCT features are beneficial, the model gains an average F1 improvement of 7.06\%. \\

\begin{figure}[]
    \centering
    \includegraphics[width=0.45\textwidth]{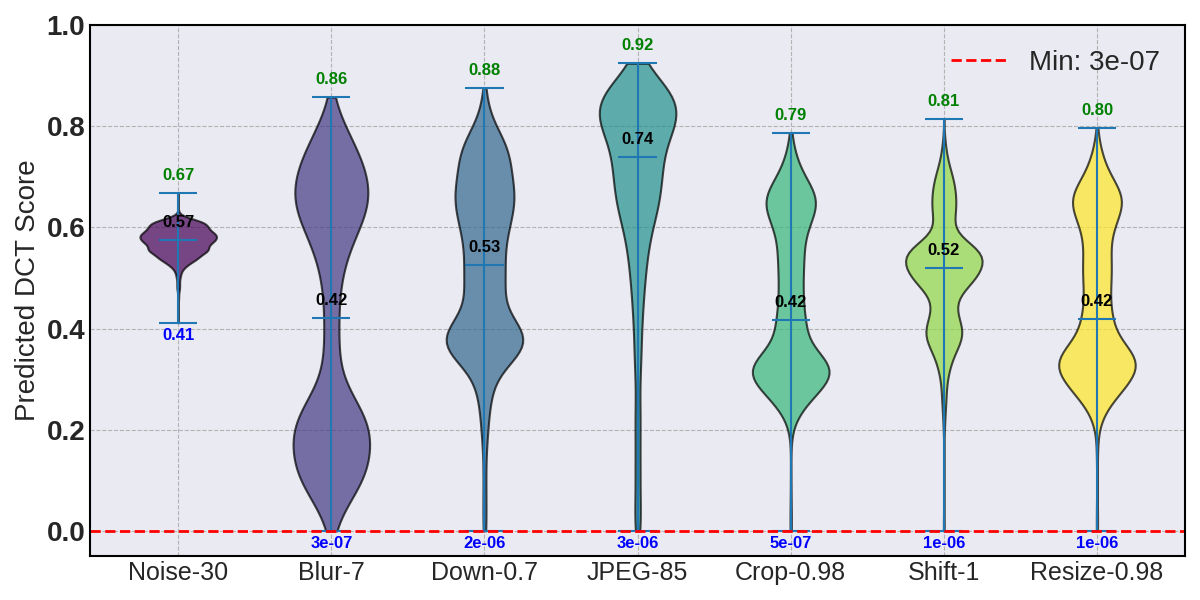}
    \vspace{-8pt}
    \caption{Distribution of predicted scores for DCT features under different distortions.}
    \label{fig:align_score_dist}
    \vspace{-5pt}
\end{figure}

\begin{figure}[t]
    \centering
    \includegraphics[width=0.48\textwidth]{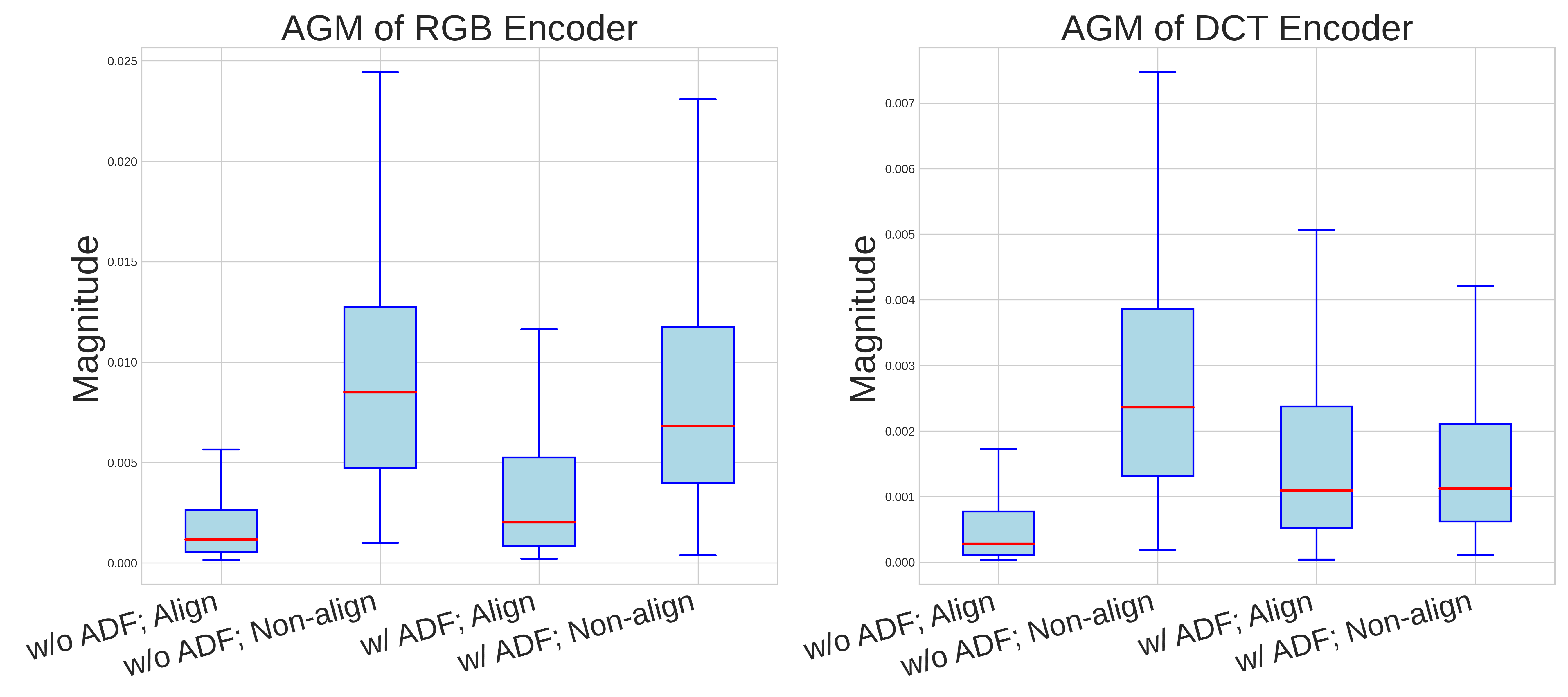}
    \vspace{-20pt}
    \caption{Average Gradient Magnitude (AGM) of $E_{\text{rgb}}$ and $E_{\text{dct}}$ for aligned/non-aligned samples, with or without ADF.}
    \label{fig:agm}
    \vspace{-15pt}
\end{figure}


\begin{figure*}[h]
    \centering
    \includegraphics[width=0.98\textwidth]{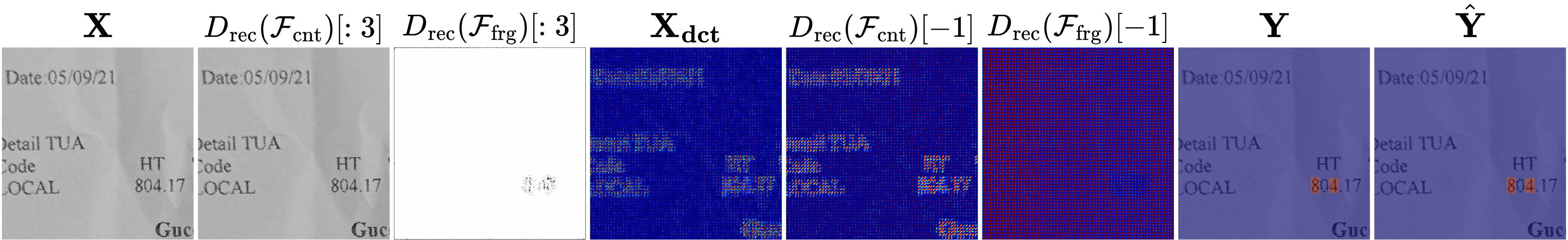}
    \vspace{-8pt}
    \caption{Visualization of HCD feature reconstructions: content features from the RGB domain \(D_{\text{rec}}(\mathcal{F}_{\text{cnt}})[:3]\) and DCT domain \(D_{\text{rec}}(\mathcal{F}_{\text{cnt}})[-1]\), and forgery features from the RGB domain \(D_{\text{rec}}(\mathcal{F}_{\text{frg}})[:3]\) and DCT domain \(D_{\text{rec}}(\mathcal{F}_{\text{frg}})[-1]\). Also shown are the original RGB image \(\mathbf{X}\), DCT coefficients \(\mathbf{X}_{\text{dct}}\), model prediction \(\hat{\mathbf{Y}}\), and ground truth \(\mathbf{Y}\).}
    \label{fig:hcd_vis}
    \vspace{-8pt}
\end{figure*}

\begin{table*}[h]
\small
\begin{center}
\renewcommand{\arraystretch}{0.8} 
\begin{tabular}{l|ccccc|ccccccc}

    \hline

    \# & DF & FL & ADF & HCD & PPE & Clean & Noise-30 & Blur-7 & JPEG-85 & Crop-0.98 & Shift-1 & AVG \\

    \hline
    
    \#1 & - & - & - & - & - & .933 & .512 & .511 & .658 & .498 & .582 & .602 \\

    \#2 & \checkmark & - & - & - & - & .929 & .653 & .577 & .703 & .560 & .652 & .661 \\

    \#3 & \checkmark & \checkmark & - & - & - & .932 & .660 & .645 & .722 & .597 & .653 & .682 \\

    \#4 & \checkmark & \checkmark & \checkmark & - & - & .922 & .570 & .690 & .791 & .684 & .658 & .719 \\

    \#5 & \checkmark & \checkmark & \checkmark & \(\mathbf{X}\) & - & .923 & .654 & .671 & .718 & .635 & .687 & .715 \\
    
    \#6 & \checkmark & \checkmark & \checkmark & \([\mathbf{X}, \mathbf{X}_{\text{dct}}]\) & - & .928 & .698 & .707 & .783 & .689 & .716 & .741 \\

    \#7 & \checkmark & \checkmark & \checkmark & - & \checkmark & .920 & .624 & .671 & .728 & .645 & .667 & .709 \\
    
    \#8 & \checkmark & \checkmark & \checkmark & \([\mathbf{X}, \mathbf{X}_{\text{dct}}]\) & \checkmark & .935 & .679 & .753 & .790 & .676 & .706 & .750 \\

    \hline
\end{tabular}
\vspace{-8pt}
\caption{Comprehensive ablation study on model settings with respect to the F1 score. DF: DCT Feature; FL: Focal Loss; ADF: Adaptive DCT Feature; HCD: Hierarchical Content Disentanglement; PPE: Pristine Prototype Estimation.}
\label{table:ablation}
\end{center}
\vspace{-25pt}
\end{table*}

\noindent\textbf{Distribution of \(\hat{s}_{\text{aln}}\) against distortions.} Fig.~\ref{fig:align_score_dist} illustrates the distribution of \(\hat{s}_{\text{aln}}\) across various distortions. We observe that distortions such as "JPEG-85" and "Down-0.7" yield relatively high \(\hat{s}_{\text{aln}}\) values. This positive correlation with \(\Delta_{\text{1vs0}}\) (Table~\ref{table:align_score}) indicates that our model effectively adjusts the DCT feature's contribution. \\

\noindent\textbf{Gradient influence of \(\hat{s}_{\text{aln}}\).} The gradient magnitude indicates feature importance~\cite{selvaraju2017grad, chen2018gradnorm}. Fig.~\ref{fig:agm} shows the average gradient magnitude (AGM) of \(E_{\text{rgb}}\) and \(E_{\text{dct}}\) for aligned and non-aligned samples. With \(\hat{s}_{\text{aln}}=1\), both features exhibit similar AGM regardless of the alignment. Conversely, applying the adaptive function \(\hat{s}_{\text{aln}} = f_{\text{asp}}(\mathbf{F}^{L}_{\text{dct}})\) yields higher AGM for \(E_{\text{dct}}\) with aligned samples and for \(E_{\text{rgb}}\) with non-aligned samples, demonstrating that our model dynamically adjusts feature importance to enhance representations in both domains. \\

\noindent\textbf{Visualization of HCD.} Fig.~\ref{fig:hcd_vis} shows the reconstructed content and forgery features in the RGB and DCT domains\footnote{To enhance visualization, the overly smooth forgery RGB image is converted to grayscale and processed with histogram equalization.}. The 4-channel output of \(D_\text{rec}\) comprises the RGB image (channels 1-3) and the DCT coefficients (last channel). The HCD module accurately recovers content and preserves expected forgery patterns in both domains. More samples are shown in the supplementary. \\

\begin{table}[h]
\small
\center
\renewcommand{\arraystretch}{0.8} 
\vspace{-15pt}
\begin{tabular}{l|cccc} 
\hline 
Dataset & \texttt{Test} & \texttt{FCD} & \texttt{SCD} & AVG \\
\hline 
TruFor & 0.392 & 0.403 & 0.285 & 0.360 \\
DTD & 0.778 & 0.825 & 0.716 & 0.773 \\
ADCD-Net (w/o HCD) & 1.014 & 0.928 & 0.868 & 0.937 \\
ADCD-Net (w/ HCD) & 1.292 & 0.934 & 1.088 & 1.105 \\
\hline 
\end{tabular}
\vspace{-8pt}
\caption{The text-BG bias \(\Delta_{S_C}\) cross models and datasets.}
\label{table:csbias}
\vspace{-10pt}
\end{table}

\noindent\textbf{Text-BG bias statistic.} To quantify the text-BG bias, we analyze the penultimate features (the outputs of the second to last layer) of different models. We compute the average cosine similarity between BG and pristine text (PT) features, \(S_C(\mathbf{\bar{f}}_{\text{bg}}, \mathbf{\bar{f}}_{\text{pt}})\), and between tampered text (TT) and PT features, \(S_C(\mathbf{\bar{f}}_{\text{tt}}, \mathbf{\bar{f}}_{\text{pt}})\). Their difference \(
\Delta S_C = S_C(\mathbf{\bar{f}}_{\text{bg}}, \mathbf{\bar{f}}_{\text{pt}}) - S_C(\mathbf{\bar{f}}_{\text{tt}}, \mathbf{\bar{f}}_{\text{pt}}) \) measures the bias, with lower values indicating greater bias. As shown in Table~\ref{table:csbias}, TruFor exhibits the most bias (\(\Delta S_C = 0.360\)). DTD with DCT traces reduces the bias by 115\%, while ADCD-Net without HCD improves \(\Delta S_C\) by 21\% due to the discriminative features produced by the FOCAL loss. Finally, ADCD-Net with HCD achieves the highest \(\Delta S_C\), showing an additional 18\% improvement and the best performance in mitigating the text-BG bias. More analysis of PPE is provided in the supplementary.

\subsection{Ablation Study} 
We conduct ablation studies to assess the impact of our proposed modules on model performance and robustness using the 3000-sample subset from 3 cross-domain test sets. These test samples are subjected to 5 types of distortion, along with the original images, to evaluate robustness. The model performance (F1 score) is given in Table~\ref{table:ablation}. We first train the backbone without any modifications (Table~\ref{table:ablation}, row \#1) as a baseline, achieving a 0.602 average F1 score across all distortions. Introducing the DCT feature and fusing it with the RGB feature (row \#2) results in a 9.9\% gain. Adding focal loss supervision (row \#3) improves performance by 13.3\%. Adaptive modulating the DCT feature (row \#4) further enhances robustness, yielding a 19.4\% gain in F1. We then investigate the impact of the content decoupling module. Recovering only the original image \(\mathbf{X}\) (row \#5) leads to a 3.5\% performance drop compared to recovering both the image and DCT coefficients (row \#6). This drop occurs because DCT coefficients, derived from the RGB image, inherit text-BG bias, where the uniform BG has more low-frequency components, while the text areas have high-frequency ones. Without the DCT coefficients recovery, this bias persists. Row \#8 shows our complete model, which achieves the highest average performance, with a 24.6\% gain over the baseline. Finally, using PPE without the content decoupling module (row \#7) causes a drop due to text-BG bias, as aggregating only BG pixels leads to inaccurate pristine cluster estimation. In turn, we observe a 5.4\% F1 score reduction compared to the complete model. 

\section{Conclusion}
We propose a specialized forgery localization model that addresses unique challenges in document images. By selectively incorporating DCT features through a learnable score, we significantly improve robustness against alignment disruptions. A hierarchical decoupling module mitigates the strong text-BG bias, enabling better detection of subtle tampering. Additionally, leveraging BG regions to estimate noise patterns enhances tampered region detection and overall model performance.

\section*{Acknowledgement}
This work was supported in part by Macau Science and Technology Development Fund under 001/2024/SK, 0022/2022/A1, and 0119/2024/RIB2; in part by Research Committee at University of Macau under MYRG-GRG2023-00058-FST-UMDF; in part by the Guangdong Basic and Applied Basic Research Foundation under Grant 2024A1515012536; by the National Natural Science Foundation of China (U2336204).

{
    \small
    \bibliographystyle{ieeenat_fullname}

}


\end{document}